\documentclass{article}

\PassOptionsToPackage{numbers, compress}{natbib}
 \usepackage[final]{nips_2016}

\usepackage[utf8]{inputenc} 
\usepackage[T1]{fontenc}    
\usepackage{hyperref}       
\usepackage{url}            
\usepackage{booktabs}       
\usepackage{amsfonts}       
\usepackage{nicefrac}       
\usepackage{microtype}      

\usepackage{amsmath}
\usepackage{amssymb}
\usepackage{amsthm}
\usepackage{algorithm}
\usepackage{algorithmic}
\usepackage{float}
\newfloat{Subroutine}{thp}{lop}
\usepackage{eqparbox} 
\usepackage{placeins}

\usepackage{setspace}
\usepackage{xspace}

\usepackage[disable]{todonotes}
\newcommand{\todod}[1]{}
\newcommand{\tododout}[1]{}
\newcommand{\todom}[1]{}
\newcommand{\todomi}[1]{}
\newcommand{\todoa}[1]{}
\newcommand{\todoaout}[1]{}

\usepackage{hyperref}

\usepackage{xcolor} 
\usepackage{wrapfig}
\usepackage{array}
\usepackage{enumitem}
\usepackage{mdframed}
\usepackage{bm}
\usepackage{amsfonts} 

\usetikzlibrary{decorations.pathreplacing,calc}
\newcommand{\tikzmark}[1]{\tikz[overlay,remember picture] \node (#1) {};}

\newcommand*{\AddNote}[4]{%
    \begin{tikzpicture}[overlay, remember picture]
        \draw [decoration={brace,amplitude=0.35em},decorate, black!70, ultra thick]
            ($(#3)!(#1.north)!($(#3)-(0,1)$)$)+(0,0.5\baselineskip) --  
            ($(#3)!(#2.south)!($(#3)-(0,1)$)$)
                node [align=left, text width=2.5cm, pos=0.5, anchor=west] {\hspace{0.15cm}#4};
    \end{tikzpicture}
}

\def\:#1{\protect \ifmmode {\mathbf{#1}} \else {\textbf{#1}} \fi}

\renewcommand{\epsilon}{\varepsilon}
\renewcommand{\Re}{\mathbb{R}}
\newcommand{\X}{\mathcal{X}}

\newcommand{\nystrom}{Nystr\"{o}m\xspace}
\newcommand{\sequentialalg}{\textsc{SQUEAK}\xspace}

\newcommand{\inkestimate}{\textsc{INK-Estimate}\xspace}

\newcommand{\updatedictop}{\textsc{Dict-Update}\xspace}

\newcommand{\shrinkop}{\textsc{Shrink}\xspace}
\newcommand{\expandop}{\textsc{Expand}\xspace}

\newcommand{\coldict}{\mathcal{I}}

\newcommand{\wt}[1]{\widetilde{#1}}
\newcommand{\wh}[1]{\widehat{#1}}

\newcommand{\wb}[1]{\overline{#1}}
\newcommand{\transp}{\mathsf{T}}
\DeclareMathOperator*{\argmin}{arg\,min}

\DeclareMathOperator*{\Tr}{Tr}

\newcommand{\normsmall}[1]{\Vert #1 \Vert}

\renewcommand{\Re}{\mathbb{R}}
\newcommand{\R}{\mathcal{R}}
\newcommand{\Real}{\mathbb{R}}

\newcommand{\kerfunc}{\mathcal{K}}
\newcommand{\kermatrix}{{\:K}}

\newcommand{\pdeff}{d_{\text{eff}}(\gamma)}
\newcommand{\adeff}[1]{\wt{d}_{\text{eff}}(#1)}
\newcommand{\atau}{\wt{\tau}}

\newcommand{\akermatrix}{\:{\wt{K}}}

\newcommand{\selmatrix}{{\:S}}

\newcommand{\dataset}{\mathcal{D}}

\newcommand{\vareps}{\varepsilon}
\renewcommand{\epsilon}{\varepsilon}
\newcommand{\bigotime}{\mathcal{O}}

\newtheorem{theorem}{Theorem}

\newtheorem{definition}{Definition}
\newtheorem{lemma}{Lemma}
\newtheorem{proposition}{Proposition}

\title{Pack only the essentials: Adaptive dictionary learning for kernel ridge regression}

\author{Daniele Calandriello   \hspace{2em}  Alessandro Lazaric \hspace{2em} Michal Valko\\ SequeL team, INRIA Lille - Nord Europe, France\\ \texttt{\small \{daniele.calandriello, alessandro.lazaric, michal.valko\}@inria.fr}}
\begin{document}
\maketitle
\vskip -1em 

\vspace{-0.3in}
\section{Introduction}\label{sec:intro}
\vspace{-0.1in}

One of the major limits of kernel ridge regression (KRR) is that for $n$ samples storing and
manipulating the kernel matrix $\kermatrix_n$ requires $\bigotime(n^2)$ space,
which becomes rapidly unfeasible for large $n$. Many solutions focus
on how to scale KRR by reducing its space (and time) complexity without
compromising the prediction accuracy. A popular approach is to construct
low-rank approximations of the kernel matrix by randomly selecting a subset of
$m$ columns from $\kermatrix_n$, thus reducing the space complexity to
$\bigotime(nm)$. These methods, often referred to as \textit{Nystr\"{o}m
approximations}, mostly differ in the distribution used to sample the columns of
$\kermatrix_n$ and the construction of low-rank approximations. Both of these
choices significantly affect the accuracy of the resulting
approximation~\citep{rudi2015less}. \citet{bach2013sharp} showed that uniform
sampling preserves the prediction accuracy of KRR (up to $\varepsilon$) only
when the number of columns $m$ is proportional to the maximum degree of freedom
of the kernel matrix. This may require sampling $\bigotime(n)$ columns in
datasets with high coherence~\citep{gittens2013revisiting}
 (i.e., a kernel matrix with weakly correlated
columns). Alternatively, \citet{alaoui2014fast} showed that sampling
columns according to their ridge leverage scores (RLS) (i.e., a measure of the
influence of a point on the regression) produces an accurate \nystrom
approximation with only a number of columns~$m$ proportional to the average
degrees of freedom of the matrix, called \textit{effective dimension}.
Unfortunately, the complexity of computing RLS is comparable to solving KRR
itself, making this approach unfeasible. However, \citet{alaoui2014fast}
proposed a fast method to compute a constant-factor approximation of the RLS
and showed that accuracy and space complexity are close to the case of sampling
with exact RLS at the cost of an extra dependency on the inverse of the minimal
eigenvalue of the kernel matrix. Unfortunately, the minimal eigenvalue can be arbitrarily small in many problems. \citet{calandriello2016analysis} addressed this
issue by processing the dataset \emph{incrementally} and updating estimates of
the ridge leverage scores, effective dimension, and \nystrom approximations
on-the-fly. Although the space complexity of the resulting algorithm
(\inkestimate) does not depend on the minimal eigenvalue anymore, it introduces
a dependency on the largest eigenvalue of $\kermatrix_n$, which in the worst
case can be as big as $n$. This can potentially reduce the advantage of the
method. In this paper we introduce \sequentialalg, a new algorithm that builds on \inkestimate,
but uses \emph{unnormalized} RLS and an improved RLS estimator. As a consequence,
the algorithm is simpler, does not need
to compute an estimate of the effective dimension for normalization,
and it achieves a space complexity that is only a constant factor worse than
sampling according to the exact RLS.


\vspace{-0.1in}
\section{Background}\label{sec:setting}
\vspace{-0.1in}

\textbf{Notation.}
We use curly capital letters $\mathcal{A}$ for collections and $|\mathcal{A}|$ for the number of entries in $\mathcal{A}$, upper-case bold letters $\:A$ for matrices and lower-case bold letters $\:a$ for vectors. We denote by $[\:A]_{ij}$ and $[\:a]_i$ the $(i,j)$ element of a matrix and $i$-th element of a vector respectively. 
We use $\:e_{n,i} \in \Re^{n}$ for the $i$-th indicator vector of dimension $n$. 
Finally, the set of the first $n$ integers is $[n] := \{1,\ldots,n\}$.

\textbf{Kernel regression.}
We consider a regression dataset $\dataset = \{(\:x_t,y_t)\}_{t=1}^n$, with input $\:x_t \in \X \subseteq \Re^d$ and output $y_t = f^\star(x_t) + \eta_t$, where $f^\star$ is an unknown target function and $\eta_t$ is a zero-mean i.i.d.\ noise. We denote by $\kerfunc: \X \times \X \rightarrow \Re$ a positive definite kernel function.
Given the first $t$ samples in~$\dataset$, the kernel matrix $\kermatrix_t \in \Re^{t\times t}$ is obtained as $[\kermatrix_t]_{ij} = \kerfunc(\:x_i, \:x_j)$ for any $i,j\in [t]$ and we denote by $\:y_t,\:f^\star_t\in\Re^t$ the vectors with components $y_i$ and $f^\star(\:x_i)$, $i\in [t]$.
Whenever a new point $\:x_{t+1}$ arrives, the kernel matrix $\kermatrix_{t+1} \in \Real^{t+1 \times t+1}$ is obtained by \emph{bordering} $\:K_t$ as
\begin{align}\label{eq:ker.bordering}
\kermatrix_{t+1} = \left[
    \begin{array}{c|c}
        \kermatrix_t & \wb{\:k}_{t+1} \\
    \hline
    \wb{\:k}_{t+1}^\transp & k_{t+1}
    \end{array}
    \right]
\end{align}
where $\wb{\:k}_{t+1} \in \Real^{t}$ is such that $[\wb{\:k}_{t+1}]_i = \kerfunc(\:x_{t+1}, \:x_i)$ for any $i\in [t]$ and $k_{t+1} = \kerfunc(\:x_{t+1}, \:x_{t+1})$.
At any time $t$, the objective of kernel regression is to find the vector $\wh{\:w}_t \in \Real^t$ that minimizes the regularized quadratic loss
\begin{align}\label{eq:kern-regr-problem}
    \wh{\:w}_t = \argmin_{\:w}\normsmall{\:y_t - \:K_t \:w}^2 + \mu \normsmall{\:w}^2 = (\:K_t + \mu \:I)^{-1} \:y_t,
\end{align}
where $\mu\in\Re$ is a regularization parameter. 
%
If $\mu$ is properly tuned, then $\wh{\:w}_t$ achieves a near-optimal risk $\R(\wh{\:w}_t) = \mathbb{E}_\eta\big[ || \:f^\star_t - \:K_t \wh{\:w}_t ||_2^2\big]$. 
Nonetheless, the computation of the final $\wh{\:w}_n$ requires $\bigotime(n^3)$ time and $\bigotime(n^2)$ space, which is infeasible for large datasets.

\textbf{Nystr\"{o}m approximation.}
A common approach to reduce the complexity is to
(randomly) select~$m$ columns of $\kermatrix_t$
according to some distribution $\:p_{t} = \{p_{t,i}\}_{i=1}^t$ and construct the dictionary
$\coldict_{t}~=~\{(i_{j},\:k_{t,i_{j}},\wt{p}_{t,i_{j}})\}_{j=1}^m$, which 
contains the set of indices $i_j \in [t]$, the corresponding 
 columns and their weights.
 Given a dictionary $\coldict_t$, the regularized Nystr\"{o}m
approximation of $\kermatrix_t$ is obtained as
\begin{align}\label{eq:nystrom}
    \wt{\kermatrix}_t = \kermatrix_t \selmatrix_t(\selmatrix_t^\transp \kermatrix_t \selmatrix_t + \gamma\:I_m)^{-1}\selmatrix_t^\transp \kermatrix_t,
\end{align}
where the selection matrix $\selmatrix_t \in \Real^{t \times m}$ is defined as
$\selmatrix_t~=~[(\wb{q}\wt{p}_{t,i_{1}})^{-1/2}\:e_{t,i_{1}}, \dots, (\wb{q}\wt{p}_{t,i_{m}})^{-1/2}\:e_{t,i_{m}}]$, $\wb{q}$ is a constant, and $\gamma$ is a regularization term (possibly different from~$\mu$).
At this point, $\wt{\kermatrix}_t$ can be used to compute $\wt{\:w}_t = (\akermatrix_t + \mu \:I_{t})^{-1} \:y_t$ efficiently using block inversion, reducing the complexity from $\bigotime(n^3)$ to $\bigotime(nm^2 + m^3)$ time
and from $\bigotime(n^2)$ to $\bigotime(nm)$ space.

\textbf{Ridge leverage scores.} The accuracy of $\wt{\kermatrix}_t$ is strictly related to the distribution $\:p_t$ used to construct the dictionary $\coldict_t$. In particular, \citet{alaoui2014fast} showed that sampling according to the $\gamma$-ridge leverage scores (RLS) of $\kermatrix_t$ leads to an accurate Nystr\"{o}m approximation.

%
\begin{definition}\label{def:exact-lev-scores}
Given $\:K_t =  \:U_t \:\Lambda_t \:U_t^\transp$, the $\gamma$-ridge leverage score (RLS) of column $i\in [t]$ is
\begin{align}\label{eq:exact-rls}
    \tau_{t,i} = \:k_{t,i}^\transp (\kermatrix_t + \gamma\:I_t)^{-1} \:e_{t,i} = \:e_{t,i}^\transp \:K_t (\kermatrix_t + \gamma\:I_t)^{-1} \:e_{t,i},
\end{align}
Furthermore, the effective dimension of the kernel is defined as $\pdeff_t = \sum_{i=1}^t \tau_{t,i}$.
\end{definition}

Similar to standard leverage
scores (i.e., $\sum_j\:[U]_{i,j}^2$), RLSs measure the importance of each point~$\:x_i$ for the
kernel regression. Furthermore, the sum of the RLSs is the effective dimension $\pdeff_t$, which measures
the intrinsic capacity of the kernel $\kermatrix_t$ when its spectrum is
soft-thresholded by a regularization $\gamma$. 
Using RLS in constructing a Nystr\"{o}m approximation leads to the following result.

\begin{proposition}[\citet{alaoui2014fast}]\label{prop:gamma.approx}
Let $\varepsilon\in[0,1]$ and $\coldict_n$ be the dictionary built with $m$ columns selected proportionally to RLSs $\{\tau_{n,i}\}$. If $m = \bigotime(\frac{1}{\varepsilon^{2}}\pdeff_n\log(\frac{n}{\delta}))$, the Nystr\"{o}m approximation $\akermatrix_n$ is a $\gamma$-approximation of $\kermatrix_t$, that is $\:0 \preceq \kermatrix_{t} - \akermatrix_{t} \preceq  \frac{\gamma}{1-\varepsilon}\kermatrix_t(\kermatrix_t + \gamma\:I)^{-1} \preceq \frac{\gamma}{1-\varepsilon}\:I$ and the risk of $\wt{\:w}_t$ is $\R(\wt{\:w}_t) \leq (1 + \frac{\gamma}{\mu}\frac{1}{1-\varepsilon})\R(\wh{\:w}_t)$.
\end{proposition}

Unfortunately, computing exact RLS requires storing $\kermatrix_n$, and has the same $\bigotime(n^2)$ space requirement as solving Eq.~\ref{eq:kern-regr-problem}. In the next section, we introduce \sequentialalg, an RLS-based incremental algorithm able to preserve the same accuracy of Prop.~\ref{prop:gamma.approx} without requiring to know the RLS in advance, and that generates a dictionary only a constant factor larger than exact RLS sampling.


\vspace{-0.1in}
\section{Incremental Nystr\"{o}m approximation with ridge leverage scores}\label{sec:setting-bis}
\vspace{-0.1in}

\sequentialalg  (Alg.~\ref{alg:sequentialalg})  builds on the \inkestimate algorithm~\cite{calandriello2016analysis} with the major algorithmic difference that the sampling probabilities are computed directly on estimates $\tau_{t,i}$ without renormalizing them by an estimate of $\pdeff_t$. \sequentialalg introduces two key elements: \textbf{1)} an improved, accurate estimator of the RLS and \textbf{2)} an incremental sampling scheme for the construction of the dictionary $\coldict_t$.

\textbf{1) Estimation of RLS.} We introduce an RLS estimator that improves
on~\cite{calandriello2016analysis}, showing that it can be
efficiently computed. At any time $t$, let $Q_t = \sum_{i} Q_{t,i}$ be
the number of columns $|\coldict_t|$
contained in the dictionary at time $t$, and $\selmatrix_t \in \Real^{t \times Q_t}$ the selection matrix
constructed so far. Let $\wb{\:S}_{t+1}~\in~\Real^{(t+1) \times (Q_t + \wb{q})}$ be constructed as
$[\:S_t, (\wb{q})^{-1/2}\:e_{t+1,t+1},\ldots, (\wb{q})^{-1/2}\:e_{t+1,t+1}]$
by adding $\wb{q}$ copies of $\:e_{t+1,t+1}$ to the selection matrix.
Denoting $\alpha = (1+\epsilon)/(1-\epsilon)$, we define the RLS estimator as
{
\newcommand{\ttfrac}{\genfrac{}{}{}2}
\begin{align}\label{eq:rls-estimator}
    \atau_{t+1,i} =& \frac{1+\epsilon}{\alpha\gamma}\left(k_{i,i} - \:k_{t+1,i}\wb{\:S} \left(\wb{\:S}^\transp\kermatrix_{t+1}\wb{\:S} + \gamma\:I\right)^{-1}\wb{\:S}^\transp\:k_{t+1,i}\right).
\end{align}
}
\vskip -1em
If $Q_t \geq \wb{q}$, then $\atau_{t+1,i}$ can be computed in $\bigotime(Q_t^3)$
time ($\bigotime(Q_t)$ to compute $\:k_{t+1,i}\wb{\:S}$ and $\bigotime(Q_t^3)$
to invert the inner matrix) and $\bigotime(Q_t^2)$ space. If $Q_t < \wb{q}$
the same applies with $\wb{q}$ replacing $Q_t$. Furthermore, we have
the following guarantee.

\begin{lemma}\label{lem:fast-rls}
Assume that the dictionary $\mathcal{I}_t$ induces a $\gamma$-approximate kernel $\akermatrix_t$.
    Then for all $i$ such that $i \in \{\mathcal{I}_{t} \cup \{t+1\}\}$,
$\atau_{t+1,i}$ computed using Eq.~\ref{eq:rls-estimator} is an $\alpha$-approximation of the RLS $\tau_{t,i}$,
that is $\tau_{t+1,i}(\gamma)/\alpha \leq \atau_{t+1,i} \leq \tau_{t+1,i}(\gamma)$.
\end{lemma}


{\setlength{\textfloatsep}{0pt}
\begin{algorithm}[t!]
\setstretch{1.05}
\begin{algorithmic}[1]
    \renewcommand{\algorithmicrequire}{\textbf{Input:}}
    \renewcommand{\algorithmicensure}{\textbf{Output:}}
    \renewcommand\algorithmiccomment[1]{%
    \hspace{2cm} \(\triangleright\)\eqparbox{COMMENT}{#1}}
    \newlength{\firstbracelen}
    \settowidth{\firstbracelen}{Add $Q_{t+1,t+1}$ copies of $(t+1, \:k_{t+1,t+1}, \wt{p}_{t+1,t+1})$ to $\mathcal{I}_{t+1}$\hspace{0.3cm}}
    \newlength{\secondbracelen}
    \settowidth{\secondbracelen}{Add $Q_{t+1,t+1}$ copies of $(t+1, \:k_{t+1,t+1}, \wt{p}_{t+1,t+1})$ to $\mathcal{I}_{t+1}$\hspace{0.3cm}\hspace{2cm}}
    \REQUIRE Dataset {$\dataset$}, regularization $\gamma, \mu$, $\wb{q}$
    \ENSURE $\akermatrix_n$, $\wt{\:w}_n$
    \STATE Initialize $\mathcal{I}_0$ as empty, $\wt{p}_{1,0} = 1$
    \FOR{$t = 0,\dots,n-1$}
        \STATE Receive new column $[\wb{\:k}_{t+1},k_{t+1}]$
        \STATE Compute $\alpha$-approximate RLS $\{\atau_{t+1,i}: i \in \mathcal{I}_{t} \cup \{t+1\}\}$, using $\coldict_t$, $[\wb{\:k}_{t+1},k_{t+1}]$, and Eq.~\ref{eq:rls-estimator}
        \STATE Set $\wt{p}_{t+1,i} = \max\left\{\min\left\{\atau_{t+1,i},\; \wt{p}_{t,i} \right\},\; \wt{p}_{t,i}/2\right\}$\\*\vspace{-0.6\baselineskip}\rule{\secondbracelen}{0.5pt}\vspace{-0.1\baselineskip}
        \STATE Initialize $\mathcal{I}_{t+1} = \emptyset$\tikzmark{top-dictupdate}\\*\vspace{-0.6\baselineskip}\rule{\firstbracelen}{0.5pt}\vspace{-0.1\baselineskip}
        \FORALL{$j \in \{1,\dots,t\}$\tikzmark{top-shrink}}
            \STATE $Q_{t,j} = |\{i=j : i \in \mathcal{I}_{t}\}|$
            \IF{$Q_{t,j} \neq 0$}
                \STATE $Q_{t+1,j} \sim \mathcal{B}(\wt{p}_{t+1,j}/\wt{p}_{t,j}, Q_{t,j})$\label{alg:tag-update-bin-sample}
                \STATE Add $Q_{t+1,j}$ copies of $(j, \:k_{t+1,j}, \wt{p}_{t+1,j})$ to $\mathcal{I}_{t+1}$.
            \ENDIF
        \ENDFOR\tikzmark{bottom-shrink}\\*\vspace{-0.6\baselineskip}\rule{\firstbracelen}{0.5pt}\vspace{-0.12\baselineskip}
        \STATE $Q_{t+1,t+1} \sim \mathcal{B}(\wt{p}_{t+1,t+1}, \wb{q})$\tikzmark{top-expand}
        \STATE  Add $Q_{t+1,t+1}$ copies of $(t+1, \:k_{t+1,t+1}, \wt{p}_{t+1,t+1})$ to $\mathcal{I}_{t+1}$\hspace{0.3cm}\tikzmark{right-shrink}\tikzmark{right-expand}\tikzmark{bottom-expand}\hspace{2cm}\tikzmark{right-dictupdate}\tikzmark{bottom-dictupdate}\\*\vspace{-0.5\baselineskip}\rule{\secondbracelen}{0.5pt}\vspace{-0.1\baselineskip}
    \ENDFOR
    \STATE Compute $\akermatrix_n$ using $\mathcal{I}_{n}$ and Eq.~\ref{eq:nystrom}
    \STATE Compute $\wt{\:w}_n$ using $\akermatrix_n$, $\:y_n$
\AddNote{top-dictupdate}{bottom-dictupdate}{right-dictupdate}{{\small \updatedictop}}
\AddNote{top-shrink}{bottom-shrink}{right-shrink}{{\small \textsc{Shrink}}}
\AddNote{top-expand}{bottom-expand}{right-expand}{{\small \expandop}}
\end{algorithmic}
\caption{The \sequentialalg algorithm}
\label{alg:sequentialalg}
\end{algorithm}
}

\textbf{2) Sequential sampling.}
At each time step $t$, \sequentialalg receives a new column $[\wb{\:k}_{t+1}, k_{t+1}]$.
This can be implemented either by having a separate algorithm that constructs each
column sequentially and streams it to \sequentialalg, or by storing just the samples (with an additional $\bigotime(td)$  space complexity)
and computing the column once.
%
Adding a new column to the matrix can either decrease
the importance of columns already observed (i.e., if they are correlated to the new column) or
leave it unchanged (i.e., if they are orthogonal) and thus the RLS evolves as $\tau_{t+1,i} \leq \tau_{t,i}$
\cite[App.~A,~Lem.~4]{calandriello2016analysis}.
In the \updatedictop loop, the dictionary is updated to reflect the change in importance of old columns (e.g., $p_{t,i} = \tau_{t,i}$ may decrease) and to add the new column proportionally to its RLS $\tau_{t+1,t+1}$.
The dictionary $\coldict_{t}$, and the new column are used to compute new approximate RLS $\atau_{t+1,i}$ as in Eq.~\ref{eq:rls-estimator}, 
which in turn define the new sampling probabilities $\wt{p}_{t+1,i}$.
The \updatedictop phase is composed of two steps.
For each index $i \in [t]$, the \shrinkop step counts the number
of copies~$Q_{t,i}$ present in $\coldict_t$, and then draws a sample from the binomial
$\mathcal{B}(\wt{p}_{t+1,i}/\wt{p}_{t,i}, Q_{t,i})$,
where taking $\wt{p}_{t+1,i} = \min\left\{\atau_{t+1,i},\; \wt{p}_{t,i} \right\}$
ensures that the binomial probability at L\ref{alg:tag-update-bin-sample} is
well defined. The more $\wt{p}_{t+1,i}$ is lower than $\wt{p}_{t,i}$, the more
$Q_{t+1,i}$ will be lower than $Q_{t,i}$. If the probability $\wt{p}_{t+1,i}$
continues to decrease over time, it is also possible that $Q_{t+1,i}$ is
decreased to zero, and column $i$ is completely dropped from the dictionary.
Intuitively, the \shrinkop step
stochastically reduces the size of the dictionary to reflect the reductions of the
RLSs.
Conversely, the \expandop step adds the new column to the dictionary with a number of copies (from 0 to $\wb{q}$) which depends on its estimated relevance $\wt{p}_{t+1,t+1}$.
Unlike in~\cite{calandriello2016analysis}, the approximate probabilities $\wt{p}_{t,i}$ are not obtained by normalizing the approximate $\wt{\tau}_{t,i}$ by an estimate of the effective dimension and thus they do not necessarily sum to one. Yet, we guarantee that $\wt{p}_{t,i} \leq p_{t,i} \leq 1$ by construction.
Note that \sequentialalg
\emph{never} estimates again the RLS of a columns dropped
from~$\mathcal{I}_{t}$.
Moreover, computing Eq.~(\ref{eq:rls-estimator}) requires only to construct
the kernel sub-matrix for samples whose indices are in $\coldict_t$.
Therefore, if we are only interested in estimating
the approximate RLS $\atau_{t,i}$ and not the regression weights $\wt{\:w}_t$,
\sequentialalg is the first RLS sampling algorithm that can operate in a single pass over the dataset
(store and access only the samples in $\coldict_t$ instead of the whole $\dataset_t$), without ever constructing the whole matrix.
 Thm.\,\ref{thm:sequential-alg-main} guarantees that \sequentialalg succeeds
in returning a $\gamma$-approximate matrix~$\akermatrix_n$ with high probability.

\begin{theorem}\label{thm:sequential-alg-main}
Let $\alpha = \left(\frac{1+\varepsilon}{1-\varepsilon}\right)$
and $\gamma > 1$. For any $0\leq \varepsilon \leq 1$,
and $0 \leq \delta \leq 1$, if we 
run Alg.\,\ref{alg:sequentialalg} with parameter
$\wb{q} =  \bigotime(\frac{\alpha}{\varepsilon^{2}}\log(\frac{n}{\delta}))$
to compute a sequence of random dictionaries $\coldict_t$ each with a random number of entries $|\coldict_t|$,
then with probability $1-\delta$, for all iterations $t \in [n]$
\vspace{-0.2\baselineskip}
\begin{itemize}[leftmargin=0.8cm]
\itemsep0pt
\item[\textbf{(1)}] The Nystr\"{o}m approximation $\akermatrix_t$ (Eq.~\ref{eq:nystrom}) associated with $\coldict_t$ is
a $\gamma$-approximation of $\kermatrix_t$.
\item[\textbf{(2)}] The number of stored columns is $|\coldict_t| = \sum_{i} Q_{t,i} \leq \bigotime(\wb{q}\pdeff_t) \leq \bigotime(\frac{\alpha}{\varepsilon^{2}}\pdeff_n \log(\frac{n}{\delta}))$.
\item[\textbf{(3)}] The solution $\wt{\:w}_t$ satisfies $\mathcal{R}(\wt{\:w}_t) \leq (1 + \frac{\gamma}{\mu}\frac{1}{1-\varepsilon})\mathcal{R}(\wh{\:w}_t)$.
\end{itemize}
\end{theorem}
As the previous theorem holds for any $t \in [n]$, \sequentialalg has any-time guarantees on its space complexity, approximation, and risk performance. In fact, \textbf{(1)} combined with
Lem.~\ref{lem:fast-rls} shows that, at all steps, $\wt{\tau}_{t,i}$ are $\alpha$-approximate RLSs
estimates. Since adding a column to $\kermatrix_t$ can only increase the effective dimension (i.e., $\pdeff_t \leq \pdeff_{t+1}$)~
\cite[App.~A,~Lem.~5]{calandriello2016analysis}, from \textbf{(2)} we see that the number of columns stored by \sequentialalg over iterations never exceeds the budget $\bigotime(\pdeff_n \log(n))$ required by sampling columns according to the exact RLS computed over the whole dataset.
Notice that this is obtained by automatically increasing the dictionary size (and space occupation)
over time to adapt to the growth in effective dimension of the data,
which does not need to be known in advance. Furthermore, if the size of the dictionary grows too large w.r.t. the memory available,
we can still terminate the algorithm knowing that the intermediate dictionary
returned is a good approximation of the part of dataset processed. 
We can
also restart the process with a larger $\gamma$, since $\pdeff_n$ is inversely
proportional to $\gamma$.
The tradeoffs of this approach are quantified by \textbf{(3)},
which shows that all solutions $\wt{\:w}_t$
incur a risk only a factor roughly $(1 + \gamma/\mu)$ away from the corresponding
exact solution $\wh{\:w}_t$. This means that choosing a small $\gamma<\mu$ allows to achieve a risk close to the exact solution for a large range of $\mu$,
at the cost of increasing the space, while larger $\gamma$ require less space but it may prevent from tuning $\mu$ optimally.
Finally, it is important to notice that even in the worst case
$\pdeff_n = n$, \sequentialalg requires only $\log(n)$ more space than storing
the whole matrix.


\vspace{-0.2cm}
\section{Discussion}\label{sec:conclusions}
\vspace{-0.2cm}

\begin{table}[t]
\setstretch{1.5}
\centering
{\small
\begin{tabular}{|c|c|c|c|c|}
\hline
& Time  & $|\coldict_n|$ (Total space = $\bigotime(n|\coldict_n|)$) & Acc. loss & Increm.\\
        \hline
        \textsc{Exact} & $n^3$ & $n$ & $1$ & N/A  \\
        \citet{bach2013sharp} & $\frac{n{d_{\text{max}}}_n^2}{\vareps} + \frac{{d_{\text{max}}}_n^3}{\vareps}$ & $\frac{d_{\text{max},n}}{\vareps}$ & $(1+4\vareps)$ & No  \\
        \citet{alaoui2014fast} & $n (|\coldict_n|)^2$ & $\left(\frac{\lambda_{\min} + n\mu\vareps}{\lambda_{\min} - n\mu\vareps}\right)\pdeff_n + \frac{\Tr(\kermatrix_n)}{\mu\varepsilon}$ & $(1+2\vareps)^2$ & No  \\
        \citet{calandriello2016analysis} & $\frac{\lambda_{\max}^2}{\gamma^2}\frac{ n^2 \pdeff_n^2}{\vareps^2}$ & $ \frac{\lambda_{\max}}{\gamma}\frac{  \pdeff_n}{\vareps^2}$ & $(1+2\vareps)^2$ & Yes  \\
            \sequentialalg & $\frac{n^2 \pdeff_n^2}{\vareps^2}$ & $ \frac{\pdeff_n}{\vareps^2}$ & $(1+2\vareps)^2$ & Yes  \\
        \textsc{RLS-sampling} & $\frac{n \pdeff_n^2}{\vareps^2}$ & $\frac{\pdeff_n}{\vareps^2}$ & $(1+2\vareps)^2$ & N/A  \\
            \hline
\end{tabular}
}\vspace{0.5\baselineskip}
\caption{{\small Comparison of \nystrom methods. $\lambda_{\max}$ and $\lambda_{\min}$ refer to largest and smallest eigenvalues of $\protect \kermatrix_n$.}}\label{fig:table-comparison}
\vspace{-1.2\baselineskip}
\end{table}


Table~\ref{fig:table-comparison} compares several \nystrom approximation methods w.r.t.\,their space complexity and risk.
For all methods, we omit $\bigotime(\log(n))$ factors.
The space complexity of uniform sampling~\cite{bach2013sharp} scales with the maximal degree
of freedom $d_{\text{max}}$. Since
$d_{\text{max}} = n \max_i \tau_{n,i} \geq \sum_{i} \tau_{n,i} = \pdeff_n$,
uniform sampling is often outperformed by RLS sampling.
While \citet{alaoui2014fast} also sample according to RLS, their two-pass
estimator is not very accurate. In particular, the first pass requires to
sample $\bigotime\left(n\mu\vareps/(\lambda_{\min} - n\mu\vareps)\right)$
columns, which quickly grows above $n^2$ when $\lambda_{\min}$ becomes small.
Finally, \citep{calandriello2016analysis} require that the maximum dictionary
size is fixed in advance, which implies some knowledge of the effective
dimensions $\pdeff_n$, and requires estimating both $\atau_{t,i}$ and
$\adeff{\gamma}_t$. In particular, this extra estimation effort causes an
additional $\lambda_{\max}/\gamma$ factor to appear in the space complexity. This
factor cannot be easily estimated, and  causes a space
complexity of $n^3$ in the worst case. 
We also include \textsc{RLS-sampling}, a fictitious algorithm that
receives the exact RLS in input, as an ideal baseline
for all RLS sampling algorithms.
From the table, we can therefore see that \sequentialalg achieves the
same space complexity (up to constant factors) as knowing the RLS in advance.
Moreover, although in this paper we only considered fixed design KRR,
$\gamma$-approximation guarantees for $\akermatrix_n$ are commonly used in similar
problems such as random design KRR, or Kernel PCA.
Finally, with a more careful analysis, we can generalize \sequentialalg and its guarantees
to the distributed setting, where multiple machines construct dictionaries in parallel on separate datasets,
and then recursively merge them to construct a dictionary for the union of the datasets.
%

\newpage
\bibliographystyle{plainnat}
\bibliography{library}

\end{document}